\pdfoutput=1
\documentclass[runningheads]{llncs}
\usepackage{graphicx}
\usepackage{comment}
\usepackage{amsmath,amssymb} % define this before the line numbering.
\usepackage{color}

\usepackage{amsmath}
\usepackage{amsfonts}
\usepackage{bm}
\usepackage{bbm}
\newcommand{\xvec}{{\bf x}}
\usepackage{caption}
\usepackage{subfigure}
\usepackage{algorithm}
\usepackage{algorithmic}
\usepackage{booktabs}

\begin{document}
% \renewcommand\thelinenumber{\color[rgb]{0.2,0.5,0.8}\normalfont\sffamily\scriptsize\arabic{linenumber}\color[rgb]{0,0,0}}
% \renewcommand\makeLineNumber {\hss\thelinenumber\ \hspace{6mm} \rlap{\hskip\textwidth\ \hspace{6.5mm}\thelinenumber}}
% \linenumbers
\pagestyle{headings}
\mainmatter
\def\ECCVSubNumber{6769}  % Insert your submission number here

\title{Dual Mixup Regularized Learning for Adversarial Domain Adaptation} % Replace with your title

% INITIAL SUBMISSION 
\begin{comment}
\titlerunning{ECCV-20 submission ID \ECCVSubNumber} 
\authorrunning{ECCV-20 submission ID \ECCVSubNumber} 
\author{Anonymous ECCV submission}
\institute{Paper ID \ECCVSubNumber}
\end{comment}
%******************

% CAMERA READY SUBMISSION
%\begin{comment}
\titlerunning{Dual Mixup Regularized Learning for Adversarial Domain Adaptation}
% If the paper title is too long for the running head, you can set
% an abbreviated paper title here
%
\author{Yuan Wu\inst{1} \and
Diana Inkpen\inst{2} \and
Ahmed El-Roby\inst{1}}
\authorrunning{Y. Wu et al.}
% First names are abbreviated in the running head.
% If there are more than two authors, 'et al.' is used.
%
\institute{Carleton University
\email{\{yuan.wu3, Ahmed.ElRoby\}@carleton.ca} \and
University of Ottawa \email{Diana.Inkpen@uottawa.ca}}
%\end{comment}
%******************
\maketitle

\begin{abstract}
 Recent advances on unsupervised domain adaptation (UDA) rely on adversarial learning to disentangle the explanatory and transferable features for domain adaptation. However, there are two issues with the existing methods. First, the discriminability of the latent space cannot be fully guaranteed without considering the class-aware information in the target domain. Second, samples from the source and target domains alone are not sufficient for domain-invariant feature extracting in the latent space. In order to alleviate the above issues, we propose a dual mixup regularized learning (DMRL) method for UDA, which not only guides the classifier in enhancing consistent predictions in-between samples, but also enriches the intrinsic structures of the latent space. The DMRL jointly conducts category and domain mixup regularizations on pixel level to improve the effectiveness of models. A series of empirical studies on four domain adaptation benchmarks demonstrate that our approach can achieve the state-of-the-art. 
\end{abstract}

%%%%%%%%%%%%%%%%%%%%%%%%%%%%%%%%%%%%%%%%%%%%%%%%%%%%%%%

\section{Introduction}

The development of deep neural networks has significantly improved the state of the arts for a wide variety of machine learning tasks, such as computer vision\cite{krizhevsky2012imagenet}, speech recognition\cite{hinton2012deep}, and reinforcement learning \cite{silver2016mastering}. However, these advancements often rely on the existence of a large amount of labeled training data. In many real-world applications, collecting sufficient labeled data is often prohibitive due to time, financial, and expertise constraints. Therefore, there is a strong motivation to train an effective predictive model which can leverage knowledge learned from a label-abundant dataset and perform well on another label-scarce domains \cite{pan2009survey}. However, due to the existence of domain shift, deep neural networks trained on one large scale labeled dataset can be weak at generalizing learned knowledge to new datasets and tasks \cite{pan2009survey}. 

%%%%%%%%%%%%%%%%%%%%%%%%%%%%%%%%%%%%%%%%%%%%%%

\begin{figure}
\tiny
    \centering
    \includegraphics[width=0.7\linewidth]{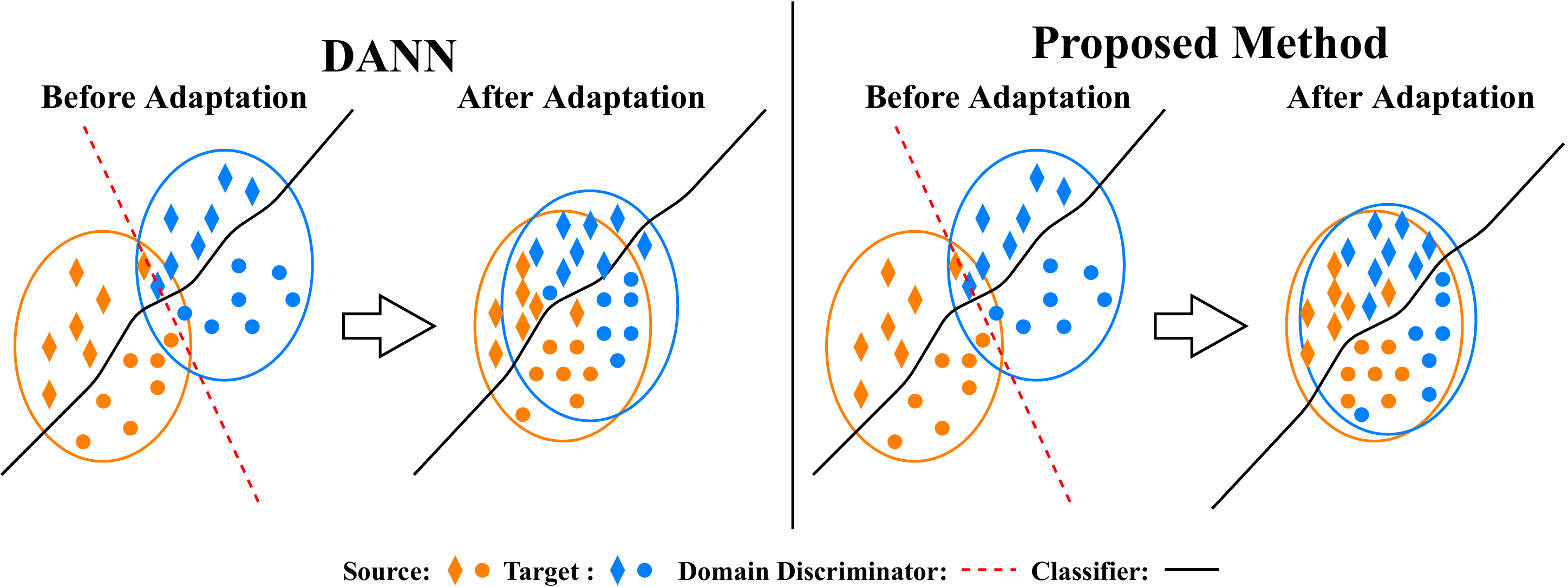}
    \caption{Comparison of DANN and the proposed method. \textbf{Left}: DANN only tries to match the feature distribution by utilizing adversarial learning; it does not consider the class-aware information in the target domain and samples from the source and target domains may not be sufficient to ensure domain-invariance of the latent space. \textbf{Right}: Our proposed method uses category mixup regularization to enforce prediction consistency in-between samples and domain mixup regularization to explore more intrinsic structures across domains, resulting in better adaptation performance.}
    \label{fig_1}
\end{figure}

%%%%%%%%%%%%%%%%%%%%%%%%%%%%%%%%%%%%%%%%%%%%%%%%%%

To address the above issue, a general strategy called domain adaptation is introduced by transferring knowledge from a label-rich domain, referred as the source domain, to a label-scarce domain, referred as the target domain \cite{ben2010theory}. Unsupervised domain adaptation addresses a more challenging scenario where there is no labeled data in the target domain. A theoretical analysis of domain adaptation is introduced by \cite{ben2010theory}, it suggests that in UDA tasks, the risk on the target domain can be bounded by the risk of a model on the source domain and the discrepancy between distributions of the two domains. Early UDA methods learned to reduce the discrepancy between domains in a shallow regime \cite{gong2012geodesic} or to re-weight source instances based on their relevance to the target domain \cite{gong2013connecting}. Later on, Maximum Mean Discrepancy (MMD) \cite{gretton2007kernel} was proposed to measure the distribution difference between source and target domains \cite{tzeng2014deep,long2015learning,long2017deep}. More recently, the UDA models are largely built on deep neural networks, and focus on learning domain-invariant features across domains by using adversarial learning \cite{ganin2015unsupervised}. Adversarial domain adaptation models can learn discriminative and domain-invariant features across domains by playing a minimax game between a feature extractor and a domain discriminator. The domain discriminator is trained to tell whether the sample comes from the source domain or target domain, while the feature extractor is learned to fool the domain discriminator. Many recent UDA methods based on adversarial learning can achieve the state-of-the-art performance \cite{saito2018maximum,long2018conditional,zou2019consensus}.

%%%%%%%%%%%%%%%%%%%%%%%%%%%%%%%%%%%%%%%%%%%%%%%%%%%%

Even though adversarial domain adaptation method has shown impressive performance for various tasks, such as image classification \cite{long2018conditional} and semantic segmentation \cite{saito2018maximum}. This approach still faces two issues: First, the adversarial domain adaptation does not take class-aware information in the target domain into consideration. Second, as we always use mini-batch stochastic gradient descent (SGD) for optimization in practice, if the batch size is small, samples from the source and target domains may not be sufficient to guarantee the domain-invariance in the latent space. Therefore, after adaptation, as shown in the left side of Figure \ref{fig_1}, the classifier may falsely align target samples of one label with samples of a different label in the source domain, which leads to inconsistent predictions. 

%%%%%%%%%%%%%%%%%%%%%%%%%%%

In this paper, we propose a dual mixup regularized learning (DMRL) method, which implements category and domain mixup regularizations on pixel level to address the aforementioned issues for unsupervised domain adaptation. Mixup has the ability of generating convex combinations of pairs of training samples and their corresponding labels. Motivated by this data augmentation technique, we propose two effective regularization mechanisms including domain-level mixup regularization and category-level mixup regularization, which play crucial roles in reducing the domain discrepancy for unsupervised domain adaptation. In particular, category mixup regularization is used to enforce consistent predictions in the latent space, which is conducted on both source and target domains, achieving a stronger discriminability of the latent space. Domain mixup regularization can reveal more mixed instances within each domain and allows the model to enrich internal feature patterns in the latent space, which can lead to a more continuous domain-invariant latent space and help match the global domain statistics across different domains. By using the two mixup-based regularization mechanisms, our model can effectively generate discriminative and domain-invariant representations. Empirical studies on four benchmarks demonstrate the performance of our approach. The contributions of our paper are summarized as follows:

%%%%%%%%%%%%%%%%%%%%%%%%%%%%%%%%%%%%

 \begin{itemize}
     \item We propose a dual mixup regularized learning method which can project the source and target domains to a common latent space, and efficiently transfer the knowledge learned from the labeled source domain to the unlabeled target domain.
     \item The proposed regularization mechanisms (category-level mixup regularization and domain-level mixup regularization) can learn discriminative and domain-invariant representations effectively to help reduce the distribution discrepancy across different domains.
     \item We empirically confirm the effectiveness of our proposed method by evaluating it on four benchmark datasets. Conducting ablation studies and parameter sensitivity analysis to validate the contributions of different components in our model and evaluate how each hyperparameter influences the performance of our method.
 \end{itemize}
 
%%%%%%%%%%%%%%%%%%%%%%%%%%%%%%%%%%%%%%%%%%%%%%%%%%%%%%%%%%%%%%%%%%%%%%%%%%%%%%%%%%%%%%%

\section{Related Work}

This work builds on two threads of research: interpolation-based regularization and domain adaptation. In this section, we briefly overview methods that are related to these two tasks.

%%%%%%%%%%%%%%%%%%%%%%%%%%%%%%%%%%%%%%%%%%%%%%%%%%%%%%%%%%%%%%%%%%%%%%%%%%%%%%%%%%%%%%

\subsection{Interpolation-based Regularization}
Interpolation-based regularization has been recently proposed for supervised learning \cite{zhang2017mixup,verma2018manifold}, it can help models to alleviate issues such as instability in adversarial training and sensitivity to adversarial samples. In particular, Mixup \cite{zhang2017mixup} is proposed to train models on virtual examples constructed as the convex combinations of pairs of inputs and labels. It can also encourage models to have a strictly linear behavior between training samples, by smoothing the models' output for a convex combination of two inputs to the convex combination of the outputs of each individual input \cite{berthelot2019mixmatch}. Moreover, Mixup can also be used to guarantee consistent predictions in the data distribution \cite{verma2019interpolation}, which can induce the separation of samples into different labels \cite{chapelle2005semi}. More recently, various variants of Mixup have been studied. A manifold extension to Mixup, Manifold-Mixup \cite{verma2018manifold}, proposes to perform interpolation in the latent space representations. \cite{berthelot2019mixmatch} exploits interpolations in the latent space generated by an autoencoder to improve performance.

%%%%%%%%%%%%%%%%%%%%%%%%%%%%%%%%%%%%%%%%%%%%%%%%%%%%%%%%%%%%%%%%%%%%%%%%%%%%%%%%

\subsection{Domain Adaptation} The main goal of domain adaptation is to transfer the knowledge learned from a label-abundant domain to a label-scarce domain. Unsupervised domain adaptation tackles a more challenging scenario where the target domain has no labeled data at all. Deep neural network based methods have been widely studied for UDA. The Deep Domain Confusion (DDC) method leverages Maximum Mean Discrepancy
(MMD) \cite{gretton2007kernel} metric in the last fully-connected layer to learn
representations that are both discriminative and transferable \cite{tzeng2014deep}. \cite{long2015learning} proposes a Deep Adaptation Network (DAN) to enhance the feature
transferability by minimizing the multi-kernel MMD
in several task specific layers. Asymmetric Tri-Training (ATT) exploits three different networks to generates pseudo-labels for target domain samples and utilizes these pseudo-labels to train the final classifier \cite{saito2017asymmetric}. Joint Adaptation Networks
(JAN) learn a transfer network by aligning the joint distributions of multiple domain-specific layers across different domains based on a joint maximum mean discrepancy criterion \cite{long2017deep}. \cite{zou2019confidence} proposes a Confidence Regularized Self-Training (CRST) framework to construct the soft pseudo-label, smoothing the one-hot pseudo-label to a conservative target distribution.

More recently, unsupervised domain adaptation methods are largely focusing on learning domain-invariant features by using adversarial training \cite{ganin2015unsupervised}. Generative Adversarial Network (GAN) is proposed in \cite{goodfellow2014generative}, which plays a minimax game between two networks: the discriminator is trained by minimizing the binary classification error of distinguishing the real images from the generated ones, while the generator is learned to generate high-quality images that are indistinguishable by the discriminator. Motivated by GAN, \cite{ganin2015unsupervised} proposes a domain adversarial neural network (DANN) that can learn discriminative and domain-invariant features by exploiting adversarial learning between a feature extractor and a domain discriminator. Many recent works have adopted the adversarial learning mechanism and achieved the state-of-the-art performance for unsupervised domain adaptation. The Adversarial Discriminative Domain Adaptation method (ADDA) uses an untied weight sharing strategy to align the feature distributions of source and target domains \cite{tzeng2017adversarial}. The Maximum Classification Discrepancy (MCD) utilizes different task-specific classifiers to learn a feature extractor that can generate category-related discriminative features \cite{saito2018maximum}. \cite{hoffman2018cycada} proposes Cycle-Consistent Adversarial Domain Adaptation (CyCADA) which implements domain adaptation at both pixel-level and feature-level by using cycle-consistent adversarial training. Multi-Adversarial Domain Adaptation (MADA) \cite{pei2018multi} can exploit multiplicative interactions between feature representations and category predictions to enforce adversarial learning. Generate to Adapt (GTA) provides an adversarial image generation approach that directly learns a joint feature space in which the distance between the source and target domains can be minimized \cite{sankaranarayanan2018generate}. Conditional Domain Adversarial Network (CDAN) conditions the domain discriminator on a multilinear map of feature representations and category predictions so as to enable discriminative alignment of multi-mode structures \cite{long2018conditional}. Consensus Adversarial Domain Adaptation (CADA) \cite{zou2019consensus} enforces the source and target encoders to achieve consensus to ensure the domain-invariance of the latent space. 

Our proposed DMRL can be regarded as an extension of this line of research by introducing both category and domain mixup regularizations on pixel level to solve complex, high dimensional unsupervised domain adaptation tasks.

%%%%%%%%%%%%%%%%%%%%%%%%%%%%%%%%%%%%%%%%%%%%%%%%%%%%%%%%%%%%%%%%%%%%%%%%%%%%%%%%%%

\section{Method}

In this paper, we consider unsupervised domain adaptation in the following setting. We have a labeled source domain $D^s=\{(x_i^s,y_i^s)\}_{i=1}^{n^s}$ with $x_i^s\in\mathcal{X}$ and $y_i^s\in\mathcal{Y}$,  
and an unlabeled target domain $D^t=\{x_i^t\}_{i=1}^{n^t}$ with $x_i^t\in\mathcal{X}$. The data in two domains are sampled from two distributions $P_S$ and $P_T$. $P_S$ and $P_T$ are assumed to be different but related (refereed as covariate shift in the literature \cite{shimodaira2000improving}). The target task is assumed to be the same with the source task. Our ultimate goal is to utilize the labeled data in the source domain to learn a predictive model $h:\mathcal{X} \mapsto \mathcal{Y}$ which can generalize well on the target domain.

%%%%%%%%%%%%%%%%%%%%%%%%%%%%%%%%%%%%%%%%%%%%%%%%%%%%%%%%%%%%%%%%%%%%%%%%%%%%%%%%

\subsection{Adversarial Domain Adaptation}

Motivated by the domain adaptation theory \cite{ben2010theory} and GANs \cite{goodfellow2014generative}, \cite{ganin2015unsupervised} proposes Domain Adversarial Neural Network (DANN), which can learn the domain-invariant features that are generalizable across domains. The standard DANN consists of three components: a feature extractor $G$, a category classifier $C$ and a domain discriminator $D$. We consider a feature extractor $G:\mathcal{X} \mapsto \mathbb{R}^m$, which maps any input instance $\xvec\in\mathcal{X}$ from the input space $\mathcal{X}$ into the latent space $G(\xvec)\in\mathbb{R}^m$; a category classifier $C:\mathbb{R}^m \mapsto \mathcal{Y}$, which transforms a feature vector in the latent space into the output label space $\mathcal{Y}$; a domain discriminator $D:\mathbb{R}^m \mapsto [0,1]$, which distinguishes the source domain data (with domain label 1) from the target domain data (with domain label 0). By training $G$ adversarially to confuse $D$, DANN can learn domain-invariant features to bridge the divergence between domains. Formally, the DANN can be formulated as:

%%%%%%%%%%%%%%%%%%%%%%%%%%%%%%%%%%%%%%%%%%%%%%%%%%%%%%%%%%%%%
\begin{figure*}
    \centering
    \includegraphics[width=0.7\columnwidth]{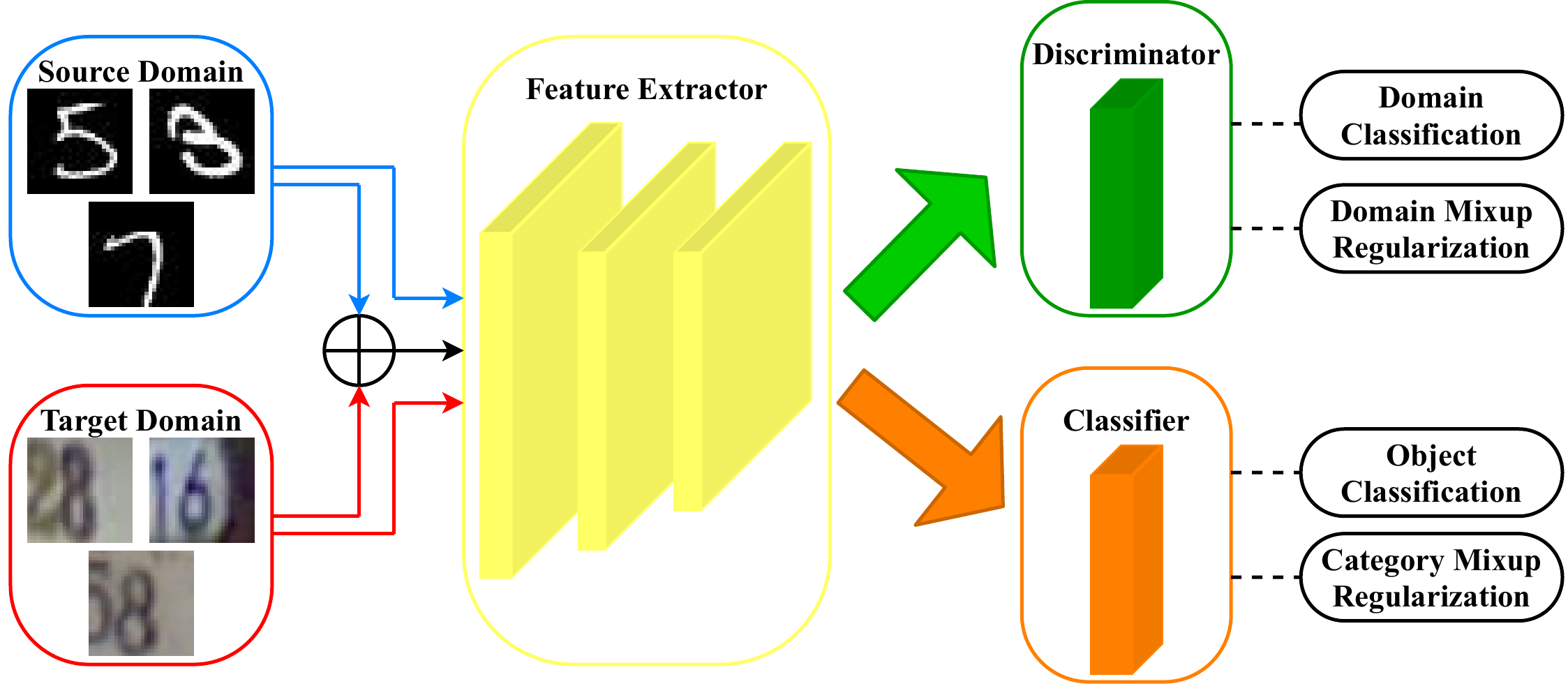}
    \caption{The architecture of the proposed dual mixup regularized learning (DMRL) method. Our DMRL consists of two mixup-based regularization mechanisms, including category-level mixup regularization and domain-level mixup regularization, which can enhance discriminability and domain-invariance of the latent space. The feature extractor $G$ aims to learn discriminative and domain-invariant features, the domain discriminator $D$ is trained to tell whether the sampled feature comes from the source domain or target domain, and the classifier $C$ is used to conduct object classification.
	}
    \label{Fig2}
\end{figure*}

%%%%%%%%%%%%%%%%%%%%%%%%%%%%%%%%%%%%%%%%%%%%%%%%%%%%%%%%%%%%

\begin{equation} \label{eq1}
    \begin{aligned}
        \min\limits_G\max\limits_D \quad \mathcal{L}_c(G,C)+\lambda_d\mathcal{L}_{Adv}(G,D)
    \end{aligned}
\end{equation}

\begin{equation} \label{eq2}
    \begin{aligned}
        \mathcal{L}_c(G,C)=\mathbb{E}_{(\xvec^s,y^s)\sim D^s}\ell(C(G(\xvec^s)),y^s)
    \end{aligned}
\end{equation}

\begin{equation} \label{eq3}
    \begin{aligned}
         \mathcal{L}_{Adv}(G,D) = \mathbb{E}_{\xvec^s\sim{D^s}}\log D(G(\xvec^s)) + \mathbb{E}_{\xvec^t\sim{D^t}}\log(1-D(G(\xvec^t)))
\end{aligned}
\end{equation}

where $\ell(\cdot,\cdot)$ is the canonical cross-entropy loss, and $\lambda_d$ is a trade-off hyperparameter. 

\subsection{Dual Mixup Regularization}

In this work, we propose a dual mixup regularized learning (DMRL) method based on adversarial domain adaptation. This method conducts category and domain mixup on pixel level. In general, Mixup performs data augmentation by constructing virtual samples with convex combinations of a pair of samples and their corresponding labels: $(\xvec_i,y_i)$ and $(\xvec_j,y_j)$:

\begin{align}
    \widetilde{\xvec}&=\mathcal{M}_{\lambda}(\xvec_i,\xvec_j)=\lambda \xvec_i+(1-\lambda)\xvec_j\\
    \widetilde{y}&=\mathcal{M}_{\lambda}(y_i,y_j)=\lambda y_i+(1-\lambda)y_j
\end{align}

where $\lambda$ is randomly sampled from a beta distribution $Beta(\alpha,\alpha)$ for $\alpha\in(0,\infty)$. By encouraging linear interpolation regularization in-between training samples, Mixup has been demonstrated effective in both supervised and semi-supervised learning \cite{zhang2017mixup,verma2019interpolation}.

%%%%%%%%%%%%%%%%%%%%%%%%%%%%%%%%%%%%%%%%%%%%%%%%%%%%%%%%%%%%%
\begin{figure*}
    \centering
    \includegraphics[width=0.7\columnwidth]{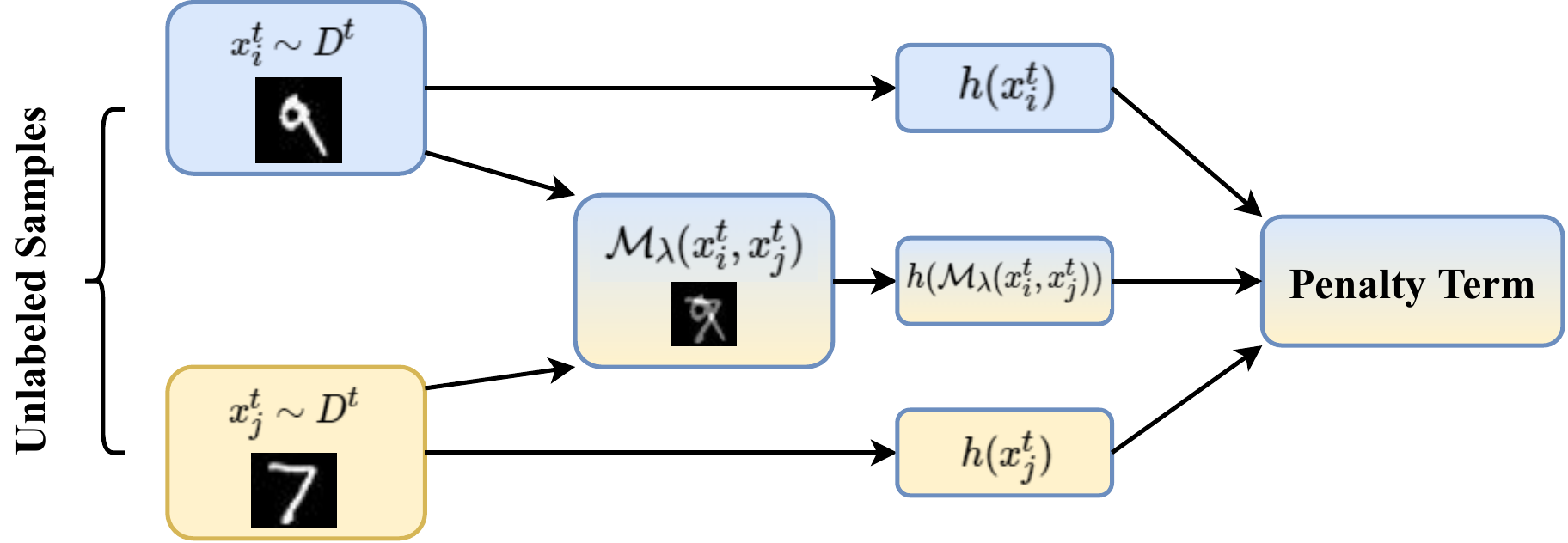}
    \caption{The framework of unlabeled category mixup regularization.
	}
    \label{Fig3}
\end{figure*}

%%%%%%%%%%%%%%%%%%%%%%%%%%%%%%%%%%%%%%%%%%%%%%%%%%%%%%%%%%%%

We largely enhance the ability of prior adversarial-learning-based domain adaptation methods by our proposed dual mixup regularized learning module, which jointly conducts category and domain mixup on pixel level to guide adversarial learning. Figure \ref{Fig2} depicts the whole framework of our proposed method. For the input, images are linearly mixed by pixel-wise addition within each individual domain. Therefore, in the input space,
there exists four kinds of samples: source samples, target samples, mixed source samples obtained by mixing two source samples, and mixed target samples obtained by mixing two target samples. After that, there exists two streams. For one stream, features of the source and target domains are used to align the global distribution statistics and conduct domain mixup regularization by the domain discriminator $D$. For the other stream, object classification and category mixup regularization are implemented by the classifier $C$. More details are provided in the following parts.

\subsubsection{Category Mixup Regularization}

The category mixup regularization consists of two components: labeled category mixup regularization and unlabeled category mixup regularization. For the source domain, since we have the labeled samples, we directly use mixed source samples and their corresponding labels to enforce prediction consistency:

\begin{align}
    \mathcal{L}_s^r(G,C)=\mathbb{E}_{(\xvec_i^s,y_i^s),(\xvec_j^s,y_j^s)\sim D^s}\ell(h(\widetilde{\xvec}^s),\widetilde{y}^s)
\end{align}

where $h$ denotes the composition of the feature extractor $G$ and the classifier $C$, $h=G\circ C$, and it can be treated as the classification function in the input space. 

For the target domain, we have no access to the label information. Therefore, mixup needs to be applied on the pseudo-labels generated by the classifier $C$. Specifically, we replace $y_i^t$ and $y_j^t$ with $h(\xvec_i^t)$ and $h(\xvec_j^t)$, which are the current predictions of $C$. Literally, $h(\xvec_i^t)$ and $h(\xvec_j^t)$ are termed as pseudo-labels of $(\xvec_i^t)$ and $(\xvec_j^t)$. Figure \ref{Fig3} presents the process of unlabeled category mixup regularization. First, we should construct convex combinations, denoted as $(\widetilde{\xvec}^t, \mathcal{M}_{\lambda}(h(\xvec_i^t), h(\xvec_j^t)))$, of pairs of samples $(\xvec_i^t, \xvec_j^t)$ and their virtual labels $(h(\xvec_i^t), h(\xvec_j^t))$. Then we conduct regularization by enforcing $h(\widetilde{\xvec}^t)$ to be consistent with $\mathcal{M}_{\lambda}(h(\xvec_i^t), h(\xvec_j^t))$ via a penalty term:

\begin{align}
    \mathcal{L}_t^r(G,C)=\mathbb{E}_{\xvec_i^t,\xvec_j^t\sim D^t}Dis(h(\widetilde{x}^t),\mathcal{M}_{\lambda}(h(\xvec_i^t),h(\xvec_j^t)))
\end{align}

where $Dis(\cdot,\cdot)$ denotes the penalty term which can punish the difference between $h(\widetilde{\xvec}^t)$ and $\mathcal{M}_{\lambda}(h(\xvec_i^t), h(\xvec_j^t))$, encouraging a linear behavior in-between training samples and $\lambda$ equals to the one used in labeled category mixup regularization. In our experiments, we use $L1$-Norm function as the penalty term. In general, Category mixup regularization can smooth the output distribution by constructing neighboring samples of the training samples and enforce prediction consistency between the neighboring and training samples, which exploits the class-aware information of the target domain in the training process, leading to performance improvement.

\subsubsection{Domain Mixup Regularization}

In practice, as we use mini-batch SGD in training, samples from the source and target domains alone are usually insufficient for global distribution alignment. Domain mixup
has the ability of generating more intermediate samples and exploring more internal structures within each domain. Linear interpolations of each domain can be generated in the input space, and the domain mixup regularization term can be defined as follows:

\begin{equation}
    \begin{aligned}
        \mathcal{L}_{Adv}^r(G,D) = \mathbb{E}_{\xvec_i^s,\xvec_j^s\sim D^s}\log D(G(\widetilde{\xvec}^s)) + \mathbb{E}_{\xvec_i^t,\xvec_j^t\sim D^t}\log (1-D(G(\widetilde{\xvec}^t)))
    \end{aligned}
\end{equation}

where $\lambda$ is the same as the one used in category mixup regularization. Contrary to prior adversarial-learning-based domain adaptation methods \cite{ganin2015unsupervised,tzeng2017adversarial,long2018conditional}, not only the samples from source and target domains, but also the mixed samples are used to align the global distribution statistics. Finally, in our proposed domain mixup regularization applied to adversarial domain adaptation, the domain-invariance of the latent space is expected to be enhanced, and with category mixup regularization, the learned representations can be more discriminative. Formally, the DMRL method can be formulated as:

\begin{equation}
    \begin{aligned}
        \min \limits_{G,C} \max \limits_D \quad \mathcal{L}_c(G,C)+\lambda_s\mathcal{L}_s^r(G,C)+\lambda_t\mathcal{L}_t^r(G,C)
        +\lambda_d\mathcal{L}_{Adv}(G,D) 
        +\lambda_r\mathcal{L}_{Adv}^r(G,D)
    \end{aligned}
\end{equation}
                                             
where $\lambda_s$, $\lambda_t$, $\lambda_d$ and $\lambda_r$ are hyperparameters for trading off different losses.

%%%%%%
\begin{algorithm}
\caption{Stochastic gradient descent training algorithm of DMRL}\label{trainingalg}
\begin{algorithmic}[1]
\STATE{\bf Input:} Source domain: $D^s$, target domain: $D^t$ and batch size: $N$.
\STATE{\bf Output:} Configurations of DMRL
\STATE{\bf Initialize} $\alpha$, $\lambda_s$, $\lambda_t$, $\lambda_d$ and $\lambda_r$
\FOR{number of training iterations}
    \STATE $(\xvec^s, y^s)$ $\leftarrow$ RANDOMSAMPLE($D^s, N$)
    \STATE $(\xvec^t)$ $\leftarrow$ RANDOMSAMPLE($D^t, N$)
    \STATE $\lambda\leftarrow$ RANDOMSAMPLE($Beta(\alpha, \alpha)$) 
    \STATE $(\widetilde{x}^s, \widetilde{y}^s)\leftarrow$ Eq. (4, 5) $\quad$ $\#$get mixed images for the source domain
    \STATE $(\widetilde{x}^t)\leftarrow$ Eq.(4) $\quad$ $\#$get mixed images for the target domain
	\STATE Calculate $l_D$ =$\lambda_d\mathcal{L}_{Adv}(G,D)$+$\lambda_r\mathcal{L}_{Adv}^r(G,D)$;\\
	Update $D$ by ascending along gradients $\nabla l_D$.\\[.2ex] 
	\STATE Calculate $loss$=$\mathcal{L}_c(G,C)$+$\lambda_s\mathcal{L}_s^r(G,C)$+$\lambda_t\mathcal{L}_t^r(G,C)$+$\lambda_d\mathcal{L}_{Adv}(G,D)$+$\lambda_r\mathcal{L}_{Adv}^r(G,D)$;\\
	Update $G$, $C$ by descending along gradients $\nabla loss$.\\[.2ex]
\ENDFOR
\end{algorithmic}
\end{algorithm}

%%%%%%%%%%%%%%%%%%%%%%%%%%%%%%%%%%%%%%%%%%%%%%%%%%%%%%%%%%%

\subsection{Training Procedure}

The training algorithm of DMRL which uses mini-batch SGD is presented in Algorithm \ref{trainingalg}. In each iteration, we first mix the samples in the input space for the source and target domains, separately. Then the mixed samples and their constitutions are used to guide the adversarial learning and conduct two mixup-based regularizations. The category mixup regularization can enforce consistent prediction constraint, while the domain mixup regularization can enrich the feature patterns in the latent space, so that the learned representation can be both discriminative and domain-invariant. $\alpha$ is a hyperparameter that controls the selection of $\lambda$. $\lambda_s$, $\lambda_t$, $\lambda_d$ and $\lambda_r$ are hyperparameters that balance different losses. In our experiments, we set $\alpha$, $\lambda_s$ and $\lambda_r$ as 0.1, 0.0001 and 0.00001. According to our parameter sensitivity analysis, the values of $\alpha$, $\lambda_s$ and $\lambda_r$ do not influence much the adaptation performance of our approach, and these hyperparameters are kept fixed in all experiments. The value of $\lambda_t$ has an influence on the adaptation performance, so $\lambda_t$ is selected via tuning on the unlabeled test data for different tasks. $\lambda_d$ is adapted according to the strategy from \cite{ganin2016domain}.

\subsection{Discussion}

For the adversarial-learning-based domain adaptation methods, both the domain label and category label play crucial roles in filling the gap between the source and target domains.  Specifically, domain labels are used to help make global distribution alignment across different domains, and category labels can enable features to be discriminative \cite{ganin2015unsupervised,tzeng2017adversarial,long2018conditional}. These two types of information can help reduce the domain discrepancy in different aspects, and complement each other for domain adaptation. However, prior adversarial domain adaptation methods suffer two limitations: (1) The lack of class-aware information of the target domain can make the extracted features less discriminative;  (2) As we use mini-batch SGD in training, samples from source and target domains do not allow models to completely explore internal feature patterns in the latent space.

In our method, we conduct two mixup-based regularizations to alleviate the above issues. First, we apply category mixup regularization on source and target domains. Specifically, for unlabeled target data, pseudo-labels are introduced. Since there are obviously false labels as target pseudo-labels, consistent prediction constraint is exploited to suppress the detrimental influence brought by the false target pseudo-labels. Moreover, category mixup regularization can smooth the output distribution of the model by encouraging the model to behave linearly in-between training samples. Then, Mixup \cite{zhang2017mixup}, as an effective data augmentation technique, is expected to provide extra mixed samples for adversarial learning. Domain mixup regularization is used to explore more internal feature patterns in the latent space, which leads to a more continuous domain-invariant latent distribution.
In summary, our proposed approach can learn discriminative and domain-invariant representations and improve the performance of different unsupervised domain adaptation tasks.

%%%%%%%%%%%%%%%%%%%%%%%%%%%%%%%%%%%%%%%%%%%%%%%%%%%%%%%%%%%%%%%%%%%%%%%%%%%%%%

\section{Experiments}

We evaluate our proposed DMRL on unsupervised domain adaptation tasks of four benchmarks, validate the effectiveness of different components in detail and investigate the influences of different hyperparameters.

\subsection{Setup}

\subsubsection{Dataset} We conducted experiments on four domain adaptation benchmarks, \textbf{Office-31}, \textbf{ImageCLEF-DA}, \textbf{VisDA-2017} and \textbf{Digits}. 
Office-31 is a benchmark domain adaptation dataset containing images belonging to 31 classes from three domains: Amazon (A) with 2,817 images, Webcam (W) with 795 images and DSLR (D) with 498 images. We evaluate all methods on six domain adaptation tasks: \textbf{A} $\rightarrow$ \textbf{W}, \textbf{D} $\rightarrow$ \textbf{W}, \textbf{W} $\rightarrow$ \textbf{D}, \textbf{A} $\rightarrow$ \textbf{D}, \textbf{D} $\rightarrow$ \textbf{A}, and \textbf{W} $\rightarrow$ \textbf{A}.

%%%%%%%%%%%%%%%%%%%%%%%%%%%%%%%%%%%%%%%%%%%%%%%%%%%%%%%%%%%%%%%%%%%%%%%%%%%%%%%%%%

ImageCLEF-DA is a benchmark dataset for ImageCLEF 2014 domain adaptation challenges, which contains 12 categories shared by three domains: Caltech-256 (C), ImageNet ILSVRC 2012 (I), and Pascal VOC 2012 (P). Each domain contains 600 images and 50 images for each category. The three domains in this dataset are of the same size, which is a good complementation of the Office-31 dataset where different domains are of different sizes. We build six domain adaptation tasks: \textbf{I} $\rightarrow$ \textbf{P}, \textbf{P} $\rightarrow$ \textbf{I}, \textbf{I} $\rightarrow$ \textbf{C}, \textbf{C} $\rightarrow$ \textbf{I}, \textbf{C} $\rightarrow$ \textbf{P}, and \textbf{P} $\rightarrow$ \textbf{C}.

%%%%%%%%%%%%%%%%%%%%%%%%%%%%%%%%%%%%%%%%%%%%%%%%%%%%%%%%%%%%%%%%%%%%%%%%%%%%%%%%%%

VisDA-2017 is a large simulation-to-real dataset, with over 280,000 images of 12 classes in the combined training, validation, and testing domains. The source images were obtained by rendering 3D models of the same object classes as in the real data from different angles and under different lighting conditions. It contains 152,397 synthetic images. The validation and test domains comprise natural images. The validation one has 55,388 images in total. We use the training domain as the source domain and validation domain as the target domain.

%%%%%%%%%%%%%%%%%%%%%%%%%%%%%%%%%%%%%%%%%%%%%%%%%%%%%%%%%%%%%%%%%%%%%%%%%%%%%%%

For Digits domain adaptation tasks, we explore three digit datasets: \textbf{MNIST}, \textbf{USPS} and \textbf{SVHN}. Each dataset contains digit images of 10 categories (0-9). We adopt the experimental settings of CyCADA \cite{hoffman2018cycada} with three domain adaptation tasks: MNIST to USPS (\textbf{MNIST}$\rightarrow$\textbf{USPS}), USPS to MNIST (\textbf{USPS}$\rightarrow$\textbf{MNIST}), and SVHN to MNIST (\textbf{SVHN}$\rightarrow$\textbf{MNIST}).

%%%%%%%%%%%%%%%%%%%%%%%%%%%%%%%%%%%%%%%%%%%%%%%%%%%%%%%%%%%%%%%%%%%%%%%%%%%%%%%%%%

\subsubsection{Comparison Methods} We compare our proposed DMRL with state-of-the-art models. For Office-31, we compare with Deep Adaptation Network (DAN) \cite{long2015learning}, Domain Adversarial Neural Network (DANN) \cite{ganin2015unsupervised}, Joint Adaptation Network (JAN) \cite{long2017deep}, Generate to Adapt (GTA) \cite{sankaranarayanan2018generate}, Multi-Adversarial Domain Adaptation (MADA)\cite{pei2018multi}, Conditional Domain Adaptation Network (CDAN) \cite{long2018conditional} and Confidence Regularized Self-Training (CRST). For ImageCLEF-DA, we compare with DAN, DANN, JAN, MADA and CDAN. For VisDA-2017, We compare with DANN, DAN, JAN, GTA,  Maximum Classifier Discrepancy (MCD) \cite{saito2018maximum} and CDAN. To further validate our method, we also conduct experiments on digit datasets, including MNIST, USPS and SVHN, we compare with DANN, Adversarial Discriminative Domain Adaptation (ADDA)\cite{tzeng2017adversarial}, Cycle-consistent Adversarial Domain Adaptation (CyCADA)\cite{hoffman2018cycada}, MCD, CDAN and Consensus Adversaria Domain Adaptation (CADA)\cite{zou2019consensus}.

%%%%%%%%%%%%%%%%%%%%%%%%%%%%%%%%%%%%%%%%%%%%%%%%%%%%%%%%%%%%%%%%%%%%%%%%%%%%%%%%%%%%%
\subsubsection{Implementation Details}

We follow standard evaluation protocols for unsupervised domain adaptation \cite{saito2018maximum,long2018conditional}. For Office-31 and ImageCLEF-DA datasets, we utilize ResNet50 \cite{he2016deep} pre-trained on ImageNet \cite{krizhevsky2012imagenet} as the backbone. For each domain adaptation task of the Office-31 and ImageCLEF-DA datasets, we report classification results of mean $\pm$ standard error over three random trials. For VisDA-2017 dataset, we follow \cite{saito2018maximum} and use ResNet101 \cite{he2016deep} pre-trained on ImageNet \cite{krizhevsky2012imagenet} as the backbone. For Digits datasets, we use a modified version of Lenet architecture as the base network, and train the models from scratch. For each backbone, we use all its layers up to the second last one as the feature extractor $G$ and replace the last full-connected layer with a task-specific fully-connected layer as the category classifier $C$. For discriminator, we use the same architecture as DANN \cite{ganin2015unsupervised}.

We adopt mini-batch SGD with momentum of 0.9 and the learning rate annealing strategy as \cite{ganin2016domain}: the learning rate is adjusted by $\eta_p=\frac{\eta_0}{(1+\theta p)^\beta}$, where $p$ denotes the process of training epochs that is normalized to be in [0,1], and we set $\eta_0=0.01$, $\theta=10$, $\beta=0.75$, which are optimized to promote convergence and low errors on the source domain. $\lambda_d$ is progressively changed from 0 to 1 by multiplying to $\frac{1-exp(-\delta p)}{1+exp(-\delta p)}$, where $\delta=10$. For all experiments, we set the hyperparameter $\alpha$ of distribution $Beta(\alpha,\alpha)$ to 0.2 as used in \cite{zhang2017mixup}. $\lambda_s$ and $\lambda_r$ are fixed as 0.0001 and 0.00001 respectively. $\lambda_t$ is chosen in the range $\{0.1, 1, 2, 5, 6, 10\}$, we select it on a per-experiment basis relying on unlabeled target data.

%%%%%%%%%%%%%%%%%%%%%%%%%%%%%%%%%%%%%%%%%%%%%%%%%%%%%%%%%%%%

\subsection{Results}

The unsupervised domain adaptation results in terms of classification accuracy in the target domain on Office-31, ImageCLEF-DA, VisDA-2017 and Digits datasets are reported in Table \ref{table1}, \ref{table2} and \ref{table3} respectively, with results of comparison methods directly cited from their original papers wherever available.

%%%%%%%%%%%%%%%%%
\setlength{\tabcolsep}{4pt}
\begin{table}
\begin{center}
\caption{Accuracy (\%) on Office-31 .}
\label{table1}
    \centering
    \resizebox{1\linewidth}{!}{
    \smallskip \begin{tabular}{ c c c c c c c c c}
        \toprule[1pt]
        Method & A$\rightarrow$W & D$\rightarrow$W & W$\rightarrow$D & A$\rightarrow$D & D$\rightarrow$A & W$\rightarrow$A & Avg \\
        \hline
        ResNet-50 \cite{he2016deep} & 68.4$\pm$0.2 & 96.7$\pm$0.1 & 99.3$\pm$0.1 & 68.9$\pm$0.2 & 62.5$\pm$0.3 & 60.7$\pm$0.3 & 76.1 \\  
        DAN \cite{long2015learning} & 80.5$\pm$0.4 & 97.1$\pm$0.2 & 99.6$\pm$0.1 & 78.6$\pm$0.2 & 63.6$\pm$0.3 & 62.8$\pm$0.2 & 80.4 \\
        DANN \cite{ganin2015unsupervised} & 82.0$\pm$0.4 & 96.9$\pm$0.2 & 99.1$\pm$0.1 & 79.7$\pm$0.4 & 68.2$\pm$0.4 & 67.4$\pm$0.5 & 82.2 \\
        JAN \cite{long2017deep} & 85.4$\pm$0.3 & 97.4$\pm$0.2 & 99.8$\pm$0.2 & 84.7$\pm$0.3 & 68.6$\pm$0.3 & 70.0$\pm$0.4 & 84.3 \\
        GTA \cite{sankaranarayanan2018generate} & 89.5$\pm$0.5 & 97.9$\pm$0.3 & 99.8$\pm$0.4 & 87.7$\pm$0.5 & 72.8$\pm$0.3 & \textbf{71.4}$\pm$0.4 & 86.5 \\
        MADA \cite{pei2018multi} & 90.0$\pm$0.1 & 97.4$\pm$0.1 & 99.6$\pm$0.1 & 87.8$\pm$0.2 & 70.3$\pm$0.3 & 66.4$\pm$0.3 & 85.2 \\
        CDAN \cite{long2018conditional} & \textbf{93.1}$\pm$0.2 & 98.2$\pm$0.2 & \textbf{100.0}$\pm$0.0 & 89.8$\pm$0.3 & 70.1$\pm$0.4 & 68.0$\pm$0.4 & 86.6 \\
        CRST \cite{zou2019confidence} & 89.4$\pm$0.7 & 98.9$\pm$0.4 & \textbf{100.0}$\pm$0.0 & 88.7$\pm$0.8 & 72.6$\pm$0.7 & 70.9$\pm$0.5 & 86.8 \\
        \textbf{DMRL (Proposed)} & 90.8$\pm$0.3 & \textbf{99.0}$\pm$0.2 & \textbf{100.0}$\pm$0.0 & \textbf{93.4}$\pm$0.5 & \textbf{73.0}$\pm$0.3 & 71.2$\pm$0.3 & \textbf{87.9} \\
        \bottomrule[1pt]
    \end{tabular}}
\end{center}
\end{table}

%%%%%%%%%%%%%%%%%%%%%%%%%%%%%%%%%%%%%%%%%%%%%%%%%%%%%%%%%%%%%

Results on the Office-31 dataset are presented in Table \ref{table1}. The results of ResNet-50 trained with only source domain data serve as the lower bound. Our proposed approach can obtain the best performance in four of six tasks: D $\rightarrow$ W, W $\rightarrow$ D, A $\rightarrow$ D, and D $\rightarrow$ A. For task W $\rightarrow$ A, DMRL can produce competitive accuracy comparing with the state-of-the-art. It is noteworthy that DMRL results in improved accuracies in two hard transfer tasks: A $\rightarrow$ D, and D $\rightarrow$ A. For two easier tasks: D $\rightarrow$ W and W $\rightarrow$ D, our approach achieves accuracy no less than 99.0\%. Moreover, our method can achieve the best average domain adaptation accuracy on this dataset. Given the fact that the number of samples per category is limited in the Office-31 dataset, these results can demonstrate that our method manages to improve the generalization ability of adversarial domain adaptation in the target domain. 

\begin{table*}
\caption{Accuracy (\%) on ImageCLEF-DA.}
\label{table2}
    \centering
    \resizebox{0.8\linewidth}{!}{
    \smallskip \begin{tabular}{ c c c c c c c c c}
        \toprule[1pt]
        Method & I$\rightarrow$P & P$\rightarrow$I & I$\rightarrow$C & C$\rightarrow$I & C$\rightarrow$P & P$\rightarrow$C & Avg \\
        \hline
        ResNet-50 \cite{he2016deep} & 74.8$\pm$0.3 & 83.9$\pm$0.1 & 91.5$\pm$0.3 & 78.0$\pm$0.2 & 65.5$\pm$0.3 & 91.2$\pm$0.3 & 80.7 \\  
        DAN \cite{long2015learning} & 74.5$\pm$0.4 & 82.2$\pm$0.2 & 92.8$\pm$0.2 & 86.3$\pm$0.4 & 69.2$\pm$0.4 & 89.8$\pm$0.4 & 82.5 \\
        DANN \cite{ganin2015unsupervised} & 75.0$\pm$0.6 & 86.0$\pm$0.3 & 96.2$\pm$0.4 & 87.0$\pm$0.5 & 74.3$\pm$0.5 & 91.5$\pm$0.6 & 85.0 \\
        JAN \cite{long2017deep} & 76.8$\pm$0.4 & 88.0$\pm$0.2 & 94.7$\pm$0.2 & 89.5$\pm$0.3 & 74.2$\pm$0.3 & 91.7$\pm$0.3 & 85.8 \\
        MADA \cite{pei2018multi} & 75.0$\pm$0.3 & 87.9$\pm$0.2 & 96.0$\pm$0.3 & 88.8$\pm$0.3 & 75.2$\pm$0.2 & 92.2$\pm$0.3 & 85.8 \\
        CDAN \cite{long2018conditional} & 76.7$\pm$0.3 & 90.6$\pm$0.3 & 97.0$\pm$0.4 & 90.5$\pm$0.4 & 74.5$\pm$0.3 & 93.5$\pm$0.4 & 87.1 \\
        \textbf{DMRL (Proposed)} & \textbf{77.3}$\pm$0.4 & \textbf{90.7}$\pm$0.3 & \textbf{97.4}$\pm$0.3 & \textbf{91.8}$\pm$0.3 & \textbf{76.0}$\pm$0.5 & \textbf{94.8}$\pm$0.3 & \textbf{88.0} \\
        \bottomrule[1pt]
    \end{tabular}}
\end{table*}
%%%%%%%%%%%%%%%%%%%%%%%%%%%%%%%%%%%%%%%%%%%%%%%%%%%%

Results on the ImageCLEF-DA dataset are shown in Table \ref{table2}, the results reveal several interesting observations: (1) Deep transfer learning methods outperform standard deep learning methods; this validates that domain shifts cannot be captured by deep networks \cite{yosinski2014transferable}. (2) The three domains in the ImageCLEF-DA dataset are more balanced than those in the Office-31 dataset. We can verify whether the performance of domain adaptation models can be improved when domain size does not change, with these more balanced domains. From Table \ref{table2}, we can see that our approach can outperform all comparison methods in all transfer tasks but with lower improvements compared to the results of the Office-31 dataset in term of the average accuracy, which validates that the domain size may cause domain shift \cite{long2017deep}. (3) our proposed DMRL method can achieve a new state-of-the-art on the ImageCLEF-DA dataset, strongly confirming the effectiveness of our method in aligning the features across domains. Moreover, for three of six tasks: C $\rightarrow$ I, C $\rightarrow$ P and P $\rightarrow$ C, our method can produce results with larger rooms of improvement, which further illustrates the effectiveness of the two mixup-based regularization mechanisms.

%%%%%%%%%%%%%%%%%%%%%%%%%

%%%%%
\begin{table*}
\caption{Accuracy (\%) on Digits and VisDA-2017.}
\label{table3}
    \centering
    \resizebox{1\linewidth}{!}{
    \smallskip \begin{tabular}{ c c c c c | c c }
    \toprule[1pt]
    Method & MNIST$\rightarrow$USPS & USPS$\rightarrow$MNIST & SVHN$\rightarrow$MNIST & Avg & Method & Synthetic$\rightarrow$Real\\
    \hline
    No Adaptation \cite{hoffman2018cycada} & 82.2 & 69.6 & 67.1 & 73.0 & ResNet-101 \cite{he2016deep} & 52.4 \\
    DANN \cite{ganin2015unsupervised} & 90.4 & 94.7 & 84.2 & 89.8 & DANN \cite{ganin2015unsupervised} & 57.4 \\ 
    ADDA \cite{tzeng2017adversarial} & 89.4 & 90.1 & 86.3 & 88.6 & DAN \cite{long2015learning} & 61.1 \\
    CyCADA \cite{hoffman2018cycada} & 95.6 & 96.5 & 90.4 & 94.2 & JAN \cite{long2017deep} & 65.7\\
    MCD \cite{saito2018maximum} & 92.1 & 90.0 & 94.2 & 92.1 & GTA \cite{sankaranarayanan2018generate} & 69.5 \\
    CDAN \cite{long2018conditional} & 93.9 & 96.9 & 88.5 & 93.1 & MCD \cite{saito2018maximum} & 71.9 \\
    CADA \cite{zou2019consensus} & \textbf{96.4} & 97.0 & 90.9 & 95.6 & CDAN \cite{long2018conditional} & 73.7 \\
    \textbf{DMRL (Proposed)} & 96.1 & \textbf{99.0} & \textbf{96.2} & \textbf{97.2} & \textbf{DMRL (Proposed)} & \textbf{75.5}\\
    \bottomrule[1pt]     
    \end{tabular}}
\end{table*}
%%%%%%%%%%%%%%%%%%%%%%%%%%%%%%%%%%%%%%%%%%%%%%%%%%%%%%%%%%%%

Positive results are also obtained on the VisDA-2017 and Digits datasets, as shown in Table \ref{table3}. For VisDA-2017, our approach achieves the highest accuracy among all compared methods, and exceeds the baseline of ResNet-101 pre-trained on ImageNet with a great margin. For the Digits datasets, our approach can gain improvements of more than 1.5\% on two tasks: USPS $\rightarrow$ MNIST and SVHN $\rightarrow$ MNIST, while for the MNIST $\rightarrow$ USPS task, we can obtain competitive results with the existing approaches. 

%%%%%%%%%%%%%%%%%%%%%%%%%%%%%%%%%%%%%%%%%%%%%%%%%%%%%%%%%%%%

\subsection{Further Analysis}

\subsubsection{Ablation Study}

We conduct ablation study with two domain adaptation tasks on the Digits dataset, MNIST$\rightarrow$USPS and USPS$\rightarrow$MNIST, to investigate the contributions of different components in DMRL. First, in order to examine the effectiveness of domain mixup (DM) and category mixup (CM), we produce two variants of DMRL: DMRL(w/o DM) and DMRL(w/o CM). Furthermore, since category mixup consists of two components: labeled category mixup (LCM) and unlabeled category mixup (UDM), DMRL(w/o LCM) and DMRL(w/o UCM) are also taken into consideration. In addition, the very basic baseline "No Adaptation", which simply trains the model on the source domain and tests on the target domain, is included in our study as well. The comparison results are presented in Table \ref{table5}. The results show that both domain mixup and category mixup can contribute to our method, which verify the effectiveness of these two mixup-based regularizations. In particular, category mixup contributes more to the model than domain mixup. When evaluating two components in category mixup, we can see that the model without unlabeled category mixup produces worse classification accuracies for both tasks, demonstrating that class-aware information in the target domain can make a significant contribution in adversarial domain adaptation. All variants produce inferior results, and the full model with two regularizations produces the best results. This validates the contribution of both category and domain mixup regularization terms. 

%%%%%%%%%%%%%%%%%%%%%%%%%%%%%%%%%%%%%%%%%%%%%%%%%%%%%%

\begin{table}
\caption{Ablation Studies.}
\label{table5}
    \centering
    \resizebox{0.5\linewidth}{!}{
    \begin{tabular}{c c c }
    \toprule[1pt]
    Method & MNIST$\rightarrow$USPS & USPS$\rightarrow$MNIST \\
    \hline
    No Adaptation\cite{hoffman2018cycada} & 82.2 & 69.6 \\
    DMRL(w/o DM) & 94.8 & 97.8\\
    DMRL(w/o CM) & 90.3 & 92.6\\
    DMRL(w/o LCM) & 95.0 & 97.9\\
    DMRL(w/o UCM) & 90.7 & 93.3\\
    DMRL(full) & 96.1 & 99.0\\
    \bottomrule[1pt]
    \end{tabular}}
\end{table}

%%%%%%%%%%%%%%%%%%%%%%%%%%%%%%%%%%%%%%%%%%%%%%%%%%%%%%%%%%%%

%%%%%%%%%%%%%%%%%%%%%%%%%%%%%%%%%%%%%%%%%%%%%%%%%%%
\begin{figure*}[htbp]
\centering
\subfigure[$\alpha$]{
\includegraphics[width=.32\columnwidth]{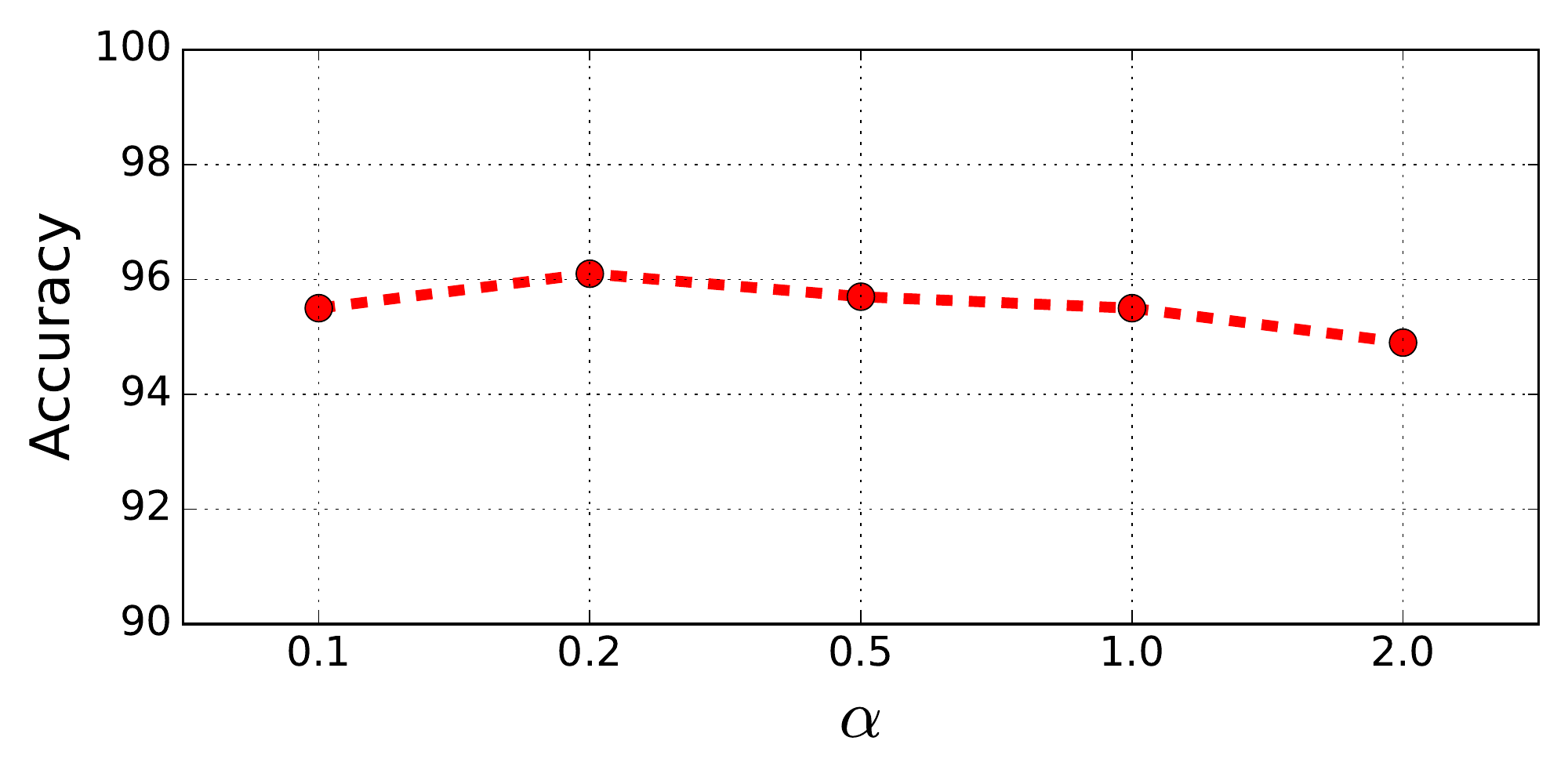}
}
\quad
\subfigure[$\lambda_s$]{
\includegraphics[width=.32\columnwidth]{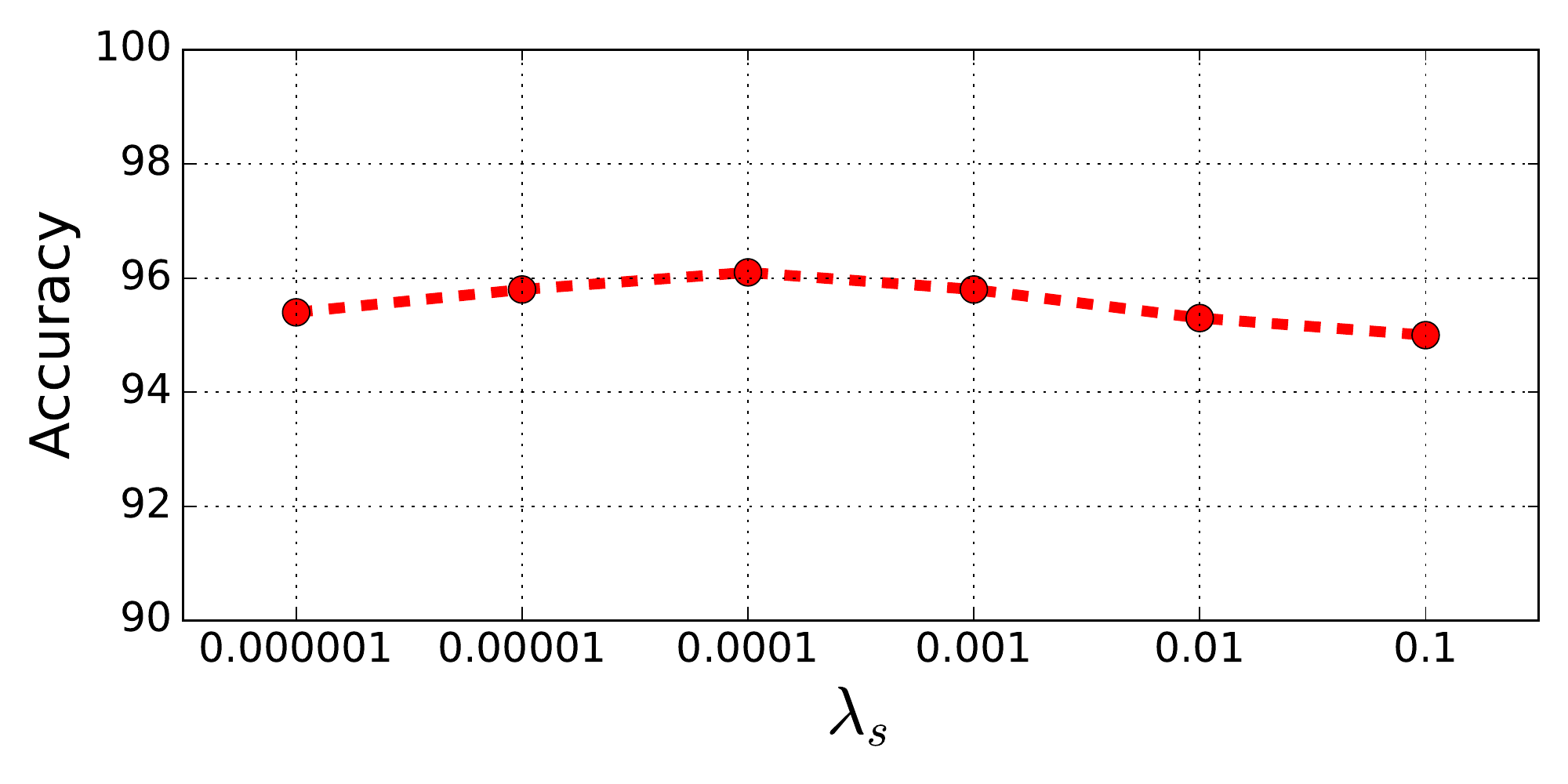}
}
\quad
\subfigure[$\lambda_r$]{
\includegraphics[width=.32\columnwidth]{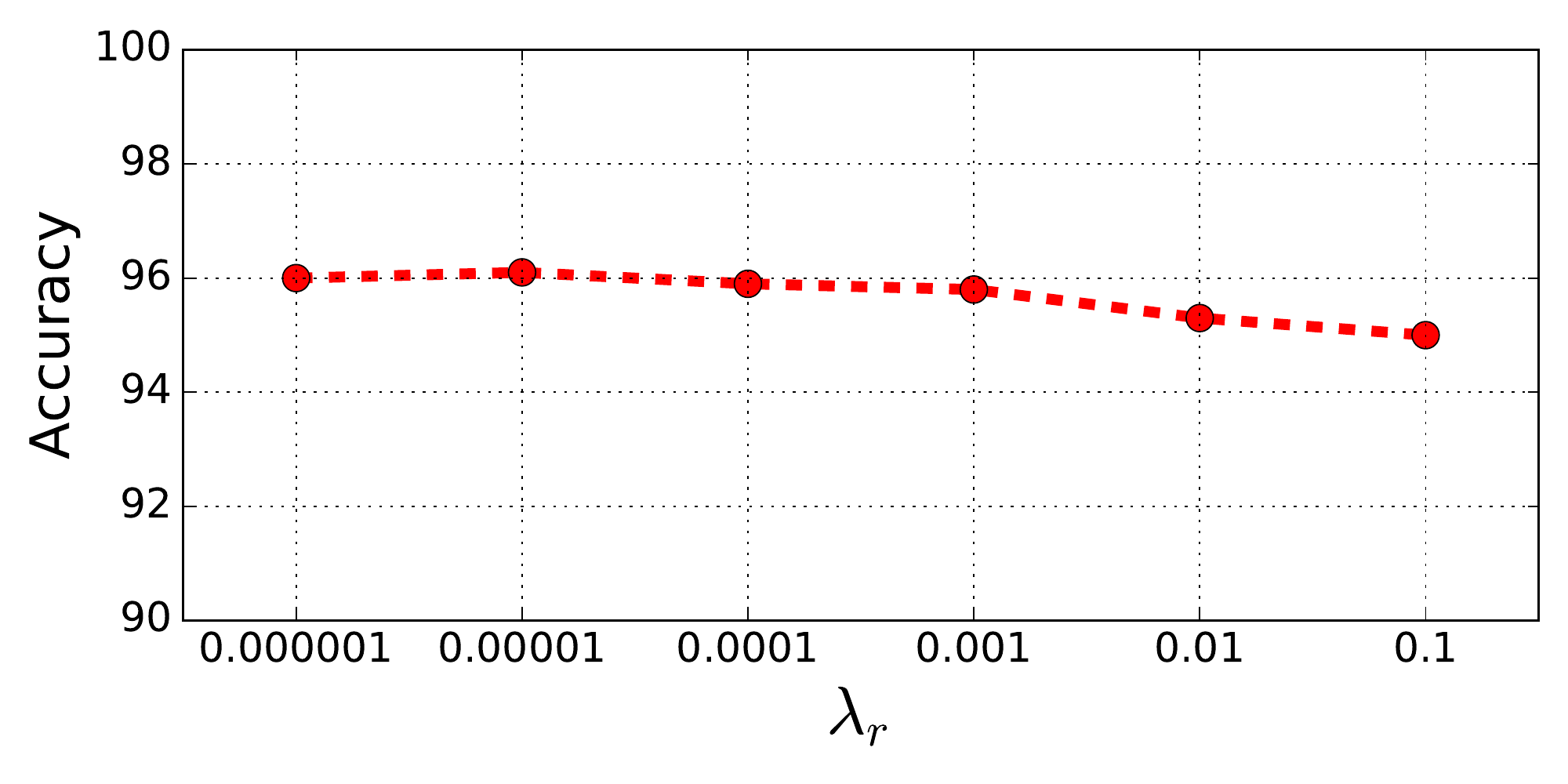}
}
\quad
\subfigure[$\lambda_t$]{
\includegraphics[width=.32\columnwidth]{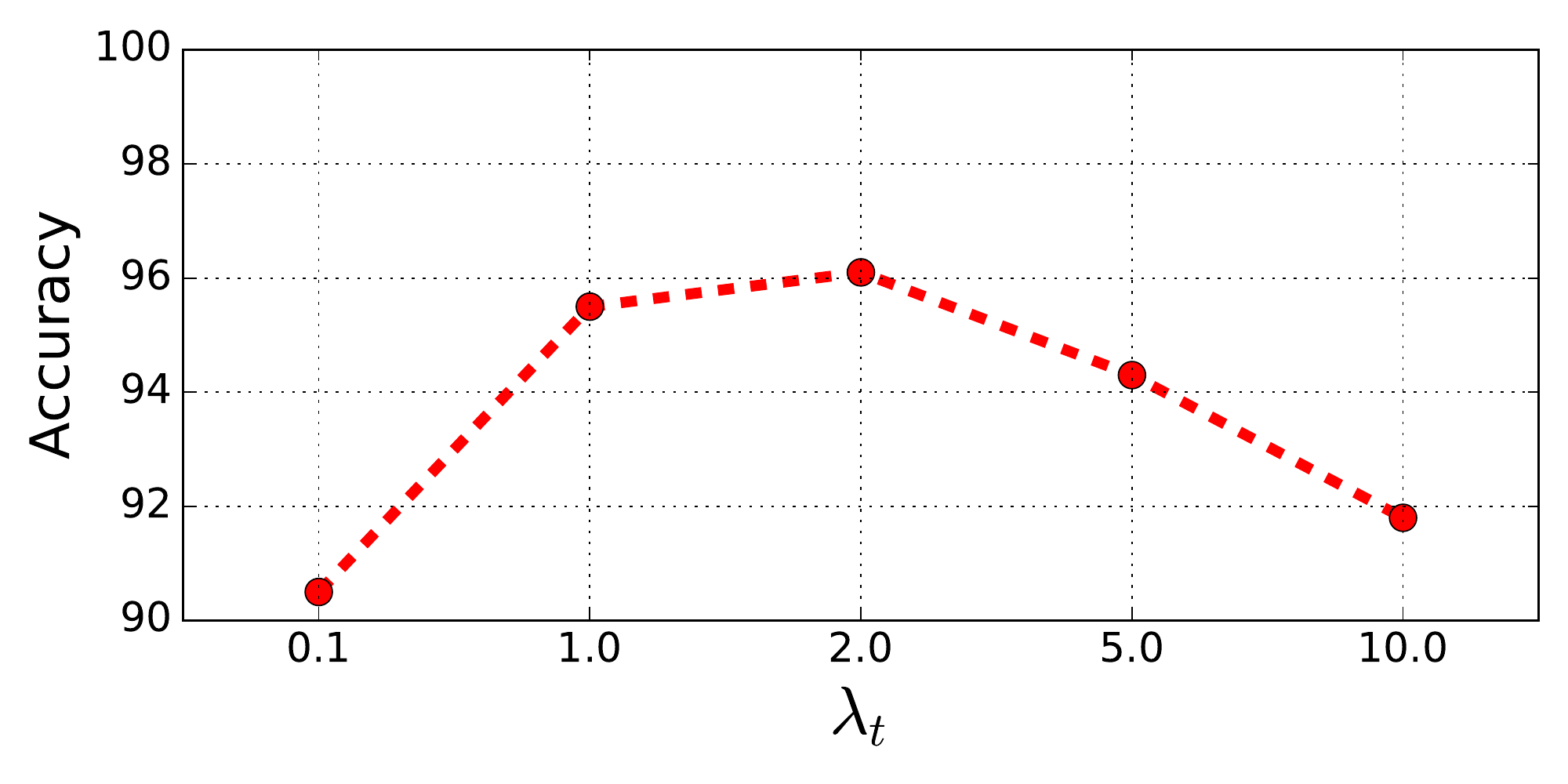}
}
\caption{Parameter sensitivity analysis}
\label{Fig4}
\end{figure*}
%%%%%%%%%%%%%%%%%%%%%%%%%%%%%%%%%%%%%%%%%%%%%%%%%%%

\subsubsection{Parameter Sensitivity Analysis}

In this section, we discuss the sensitivity of our approach to the values of the hyperparameters $\alpha$, $\lambda_s$, $\lambda_r$ and $\lambda_t$. $\lambda_s$, $\lambda_r$ and $\lambda_t$ are used to trade-off among losses, and $\alpha$ constrains the selection of $\lambda$ when conducting Mixup. We evaluate these hyperparameters on the Digits dataset, especially, the MNIST$\rightarrow$USPS task. When evaluating one hyperparameter, the others are fixed to their default values (e.g., $\alpha=2$, $\lambda_s=0.0001$, $\lambda_r=0.00001$ and $\lambda_t=2$). $\alpha$ is tested in the range $\{0.1, 0.2, 0.5, 1.0, 2.0\}$, $\lambda_s$ and $\lambda_r$ are explored in the range $\{0.000001, 0.00001, 0.0001, 0.001, 0.01, 0.1\}$, and $\lambda_t$ is evaluated in the range $\{0.1, 1, 2, 5, 6, 10\}$. The experimental results are reported in Figure \ref{Fig4}.

From Figure \ref{Fig4}, it can be observed that the domain adaptation performance is not sensitive to the hyperparameters $\alpha$, $\lambda_s$ and $\lambda_r$. Consequently, we can set $\alpha$, $\lambda_s$ and $\lambda_r$ as 0.2, 0.0001 and 0.00001 in all experiments. In addition, with the increase of $\lambda_t$, the accuracy increases dramatically and reaches the best value at $\lambda_t=2.0$, then it decreases rapidly. The parameter sensitivity analysis illustrates that a properly selected $\lambda_t$ can effectively improve the performance. 

%%%%%%%%%%%%%%%%%%%%%%%%

\section{Conclusion}

In this paper, we propose a dual mixup regularized learning (DMRL) framework for adversarial domain adaptation. By conducting category and domain mixup on pixel level, the DMRL cannot only guide the classifier in enhancing consistent predictions in-between samples, which can help avoid mismatches and enforce a stronger discriminability of the latent space, but also explore more internal structures in the latent space, which leads to a more continuous latent space. These two mixup-based regularizations can enhance and complement each other to learn discriminative and domain-invariant representations for target task. The experiments demonstrate that the proposed DMRL can effectively gain performance improvements on unsupervised domain adaptation tasks.

%%%%%%%%%%%%%%%%%%%%%%%%%%%%%%%%%%%%%%%%%%%%%%%%%%%%%%%%%%%%%%%%%%%%%%%%%%%%%%%%%%

% ---- Bibliography ----
%
% BibTeX users should specify bibliography style 'splncs04'.
% References will then be sorted and formatted in the correct style.
%
\bibliographystyle{splncs04}
\bibliography{egbib}
\end{document}